\documentclass{article}

\usepackage{arxiv}

\usepackage[utf8]{inputenc} 
\usepackage[T1]{fontenc}    
\usepackage{hyperref}       
\usepackage{url}            
\usepackage{booktabs}       
\usepackage{amsfonts}       
\usepackage{nicefrac}       
\usepackage{microtype}      
\usepackage{lipsum}
\usepackage{graphicx}
\usepackage{multicol}
\usepackage{tabularx}
\usepackage{multirow}
\usepackage{float}
\graphicspath{ {./media/} }

\title{Improving RAG Retrieval via Propositional Content Extraction: a Speech Act Theory Approach}

\author{
 João Alberto de Oliveira Lima \\
  University of Brasília; Federal Senate of Brazil\\
  \texttt{joaolima@senado.leg.br} 
}

\begin{document}
\maketitle
\begin{abstract}
When users formulate queries, they often include not only the information they seek, but also pragmatic markers such as interrogative phrasing or polite requests. Although these speech act indicators communicate the user's intent—whether it is asking a question, making a request, or stating a fact—they do not necessarily add to the core informational content of the query itself. This paper investigates whether extracting the underlying propositional content from user utterances—essentially stripping away the linguistic markers of intent—can improve retrieval quality in Retrieval-Augmented Generation (RAG) systems. Drawing upon foundational insights from speech act theory, we propose a practical method for automatically transforming queries into their propositional equivalents before embedding. To assess the efficacy of this approach, we conducted a experimental study involving 63 user queries related to a Brazilian telecommunications news corpus with precomputed semantic embeddings. Results demonstrate clear improvements in semantic similarity between query embeddings and document embeddings at top ranks, confirming that queries stripped of speech act indicators more effectively retrieve relevant content. 
\end{abstract}


\begin{multicols}{2}\raggedcolumns

\section{Introduction}

Understanding how people perform actions through language—commonly referred to as \textit{speech acts}—is essential for both effective human communication and computational linguistics. Originally in philosophical discussions by authors such as J. L. Austin \cite{austin1962}, and subsequently expanded upon by John Searle \cite{searle1969}, speech act theory introduces a critical distinction between the propositional content (the factual information or assertion) and the illocutionary force (the intent or communicative function behind the utterance). Consider the sentences: \textit{``Is the door shut?''}, \textit{``Shut the door!''},  \textit{``The door is shut.''}, and \textit{``Would that the door were shut!''}. Despite differing clearly in their intent—as a question, a command, a statement or a hope respectively—the underlying proposition remains consistent: the door being shut \cite{searle1979}. Thus, speech act theory conceptualizes communication as comprising a propositional content \textit{p} accompanied by an illocutionary force \textit{F}, typically represented in the notation \textit{F(p)}.

This distinction proves particularly relevant in fields like natural language processing (NLP) and information retrieval, especially with recent advances in Retrieval-Augmented Generation (RAG). RAG systems integrate language generation capabilities with retrieval mechanisms that fetch relevant textual information to ground generated responses in factual accuracy \cite{lewis2020}. These systems commonly employ semantic embeddings—vector representations encoding the meaning of sentences or documents—to perform retrieval tasks effectively. However, challenges arise when user queries differ stylistically or pragmatically from the format of the stored textual content. For instance, knowledge bases are typically formulated in assertive or declarative statements, while user inquiries often manifest as questions, requests, or commands. Consequently, embeddings generated for user queries and database statements, even when addressing identical factual content, may differ substantially, leading to retrieval mismatches \cite{karpukhin2020}. Such embedding discrepancies can hinder retrieval accuracy significantly. To illustrate, consider a user's query like \textit{``What is the capital of France?''} Compared to a stored factual statement such as \textit{``Paris is the capital of France.''}, embedding models might produce vectors that are not sufficiently similar, due primarily to the interrogative structure and additional function words present in the query. As highlighted in recent studies, this semantic gap between question-style queries and declarative knowledge bases limits retrieval effectiveness, causing relevant content to be overlooked during retrieval \cite{gao2022, karpukhin2020}. Consequently, one promising solution involves systematically converting user queries into forms that emphasize their core propositional content, effectively neutralizing illocutionary indicators such as question formats or polite requests.

\begin{figure}[H]
    \centering
    \includegraphics[width=.8\textwidth]{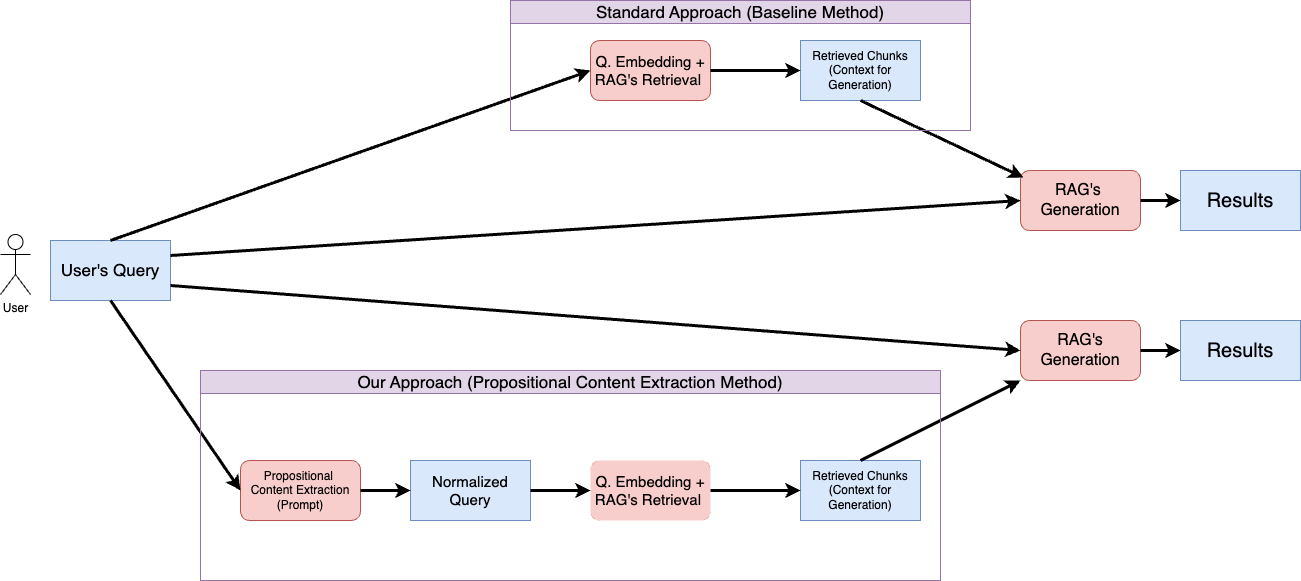}
    \caption{Comparison between the standard retrieval approach and our propositional content extraction method.}
    \label{fig:method}
\end{figure}

Figure \ref{fig:method} illustrates the methodological comparison between the standard embedding-based retrieval approach commonly used in RAG systems and our proposed method of propositional content extraction. As depicted, our approach systematically transforms user queries into normalized statements by explicitly removing illocutionary markers before embedding and retrieval.

\end{multicols}

\begin{multicols}{2}\raggedcolumns

A practical way to address this embedding mismatch involves normalizing queries to statements that represent their propositional essence. By removing linguistic markers associated with questions or requests, such as interrogative constructions (e.g., ``who,'' ``what,'' ``where'') or politeness modifiers (e.g., ``please,'' ``could you''), we obtain simpler query formulations that align better with the content stored in knowledge bases. The effectiveness of this strategy is supported by linguistic theory, which identifies the variability of speech act forms around a consistent propositional core \cite{austin1962, searle1979}.

Our study investigates this normalization approach within the context of a RAG framework. We utilize speech act theory as a theoretical foundation, drawing from Austin's \cite{austin1962} initial conceptualization of performative utterances and Searle's \cite{searle1969} subsequent refinements of illocutionary acts. Searle categorizes speech acts into assertives, directives, commissives, expressives, and declaratives—each differing in their intended effect on the listener \cite{searle1979}. Typically, user queries act as directives or interrogatives, requesting information, whereas stored documents predominantly utilize assertive forms, presenting information as facts or statements. By converting queries from directive or interrogative forms to assertive statements, we hypothesize that the embeddings generated will better match those of the stored content, facilitating more accurate retrieval.

To empirically examine this hypothesis, we introduce a method to automatically extract propositional content from user queries, guided by established linguistic principles. Our experiments involve evaluating this method in a realistic setting, applying it to a specialized domain of Brazilian telecommunications news. By measuring the performance of this approach against conventional retrieval strategies within the RAG context, we seek to demonstrate the practical benefits of integrating classical linguistic insights into modern retrieval methods.

\section{Related Work}

\subsection{Speech Act Theory and Computational Applications}

Speech act theory emerged from philosophical explorations into how language is used to perform actions, most notably through the influential work of J. L. Austin \cite{austin1962}. Austin introduced the concept that speech is not limited merely to conveying information but actively executes actions. He distinguished three types of speech acts: \textit{locutionary acts}, referring to the actual utterance and its literal meaning; \textit{illocutionary acts}, representing the intended communicative purpose such as asking, ordering, or promising; and \textit{perlocutionary acts}, relating to the impact or effect the utterance has on the listener \cite{austin1962}. John Searle expanded Austin’s foundational ideas by formalizing speech acts through the representation , where \textit{F} denotes the illocutionary force and \textit{p} symbolizes the propositional content. Searle also identified linguistic markers—known as illocutionary force indicating devices (IFIDs)—that explicitly communicate the speaker's intent \cite{searle1969}.

Building upon this theoretical framework, Searle later categorized speech acts into five distinct groups: \textit{assertives}, which commit the speaker to the truthfulness of a statement; \textit{directives}, used by speakers to prompt an action from the listener; \textit{commissives}, expressing commitments made by the speaker to future actions; \textit{expressives}, conveying psychological states or emotions; and \textit{declaratives}, which enact a change in reality merely by virtue of being spoken (e.g., officiating a marriage or opening a meeting) \cite{searle1979}. Further refinement of speech act logic by Daniel Vanderveken emphasized that various speech acts, despite differing superficially, frequently share identical propositional content, emphasizing the significance of isolating propositional meaning in computational contexts \cite{SearleVanderveken1985}.

Within computational linguistics, speech act theory has seen various applications, especially in dialogue systems and intent classification tasks. Early computational frameworks frequently involved labeling utterances with speech act or dialogue act annotations to guide system responses appropriately. The Dialogue Act Markup in Several Layers (DAMSL), for instance, gained popularity in the 1990s and 2000s, notably applied to conversational datasets like the Switchboard corpus. More recent developments leverage deep learning methods for speech act identification, improving the understanding of interactive language in structured contexts such as customer service conversations and informal dialogues \cite{Stolcke2000}. Despite advancements in computational models, a thorough integration of speech act theory into NLP remains somewhat limited, overshadowed by related areas such as sentiment analysis or intent detection. Recent surveys indicate that explicitly employing speech act classifications and pragmatics in NLP remains a promising yet underdeveloped area of research \cite{Jurafsky2009}.

\subsection{Retrieval-Augmented Generation (RAG) and Vector Embeddings}

 RAG has become an influential approach for tasks requiring knowledge-intensive natural language processing, particularly in open-domain question answering \cite{lewis2020}. The foundational work by Lewis et al. \cite{lewis2020} demonstrated that combining dense retrieval systems with generative models significantly improved accuracy in generating factually correct responses. In such setups, a dense embedding index of textual content, created using transformer-based encoders, supports efficient retrieval by measuring the semantic similarity between the query and stored documents.

At the core of RAG systems lies the embedding model, usually based on transformer architectures like BERT, which encode queries and documents into dense vector representations. Dense Passage Retrieval (DPR), introduced by Karpukhin et al. \cite{karpukhin2020}, exemplifies this approach by training dual-encoder models to produce embeddings that closely align queries with relevant document passages. The key idea behind DPR is to use contrastive learning methods, ensuring queries are embedded in close proximity to their corresponding answer-containing passages while distancing unrelated texts. However, despite extensive training, discrepancies between the query and document embedding spaces can still occur, particularly if the query formulation significantly differs from the format of the indexed documents. These differences often result in reduced retrieval effectiveness, especially when generic embeddings or pre-trained models without fine-tuning are employed.

Recent advancements have aimed at reducing these embedding mismatches through innovative strategies. One such method is the Hypothetical Document Embedding (HyDE), proposed by Gao et al. \cite{gao2022}. In HyDE, an instruction-driven language model like GPT-3 generates a plausible but hypothetical document in response to a query, transforming the original query into a form more akin to a assertive statement. Embedding this hypothetical text, rather than the original question, results in vectors more closely aligned with document embeddings, thereby improving retrieval accuracy. This generative step effectively addresses the inherent mismatch between interrogative queries and assertive knowledge statements, resulting in enhanced zero-shot retrieval capabilities. By reframing the query as a factual-like representation, HyDE achieves substantial performance gains without requiring additional annotated data for training.

Another complementary approach involves generating synthetic questions directly from document content, effectively converting the indexed information into query-like representations. Ma et al. \cite{Ma2021} explored this idea by generating potential queries from passages within the corpus, indexing these synthetic questions alongside the original texts. This allows retrieval methods to directly compare user queries against a set of pre-generated questions rather than against raw document content. Matching queries to similar queries can potentially alleviate the semantic mismatch encountered when aligning interrogative queries with assertive texts. However, this method necessitates additional processing steps to create and validate the synthetic questions, increasing complexity and potentially introducing noise or irrelevant queries into the dataset.

Our current study adopts an approach that conceptually parallels these generative methods but leverages linguistic simplification rather than generating new textual content. Instead of using language models to fabricate hypothetical answers or synthetic questions—which can risk inaccuracies or demand substantial computational resources—we propose systematically simplifying user queries to their essential propositional content. This approach entails removing linguistic markers such as interrogative syntax, politeness terms, and performative verbs, resulting in queries closer in style to declarative statements. By avoiding generative methods, our technique reduces computational overhead and the risk of introducing false or misleading information.

Embedding-based semantic search techniques, foundational to contemporary retrieval systems, offer significant improvements over traditional keyword-based search by capturing semantic similarities rather than relying solely on exact lexical matches \cite{Reimers2019}. Nevertheless, the performance of these semantic methods critically depends on embedding models accurately encoding relevant aspects of meaning. Query reformulation strategies, such as those proposed in our work, aim to refine query embeddings to better align with the semantic representations of indexed documents. This reflects long-standing practices in information retrieval, where simplifying or reformulating user queries often enhances search outcomes.

\section{Methodology}

\subsection{Propositional Content Extraction from Queries}

Our method centers on transforming user queries into simplified statements that preserve the core propositional content while removing linguistic markers indicative of illocutionary force. This transformation process, guided by principles from speech act theory, utilizes GPT-4's advanced linguistic understanding to clarify the user's true informational intent. To accomplish this, we designed a structured prompt (Appendix) that instructs GPT-4 to systematically remove pragmatic markers while preserving core propositions. Table \ref{tab:speechact} provides examples illustrating how different speech acts are processed into their propositional equivalents.

It is important to clarify our rationale for including additional speech act categories beyond those originally defined by Searle. While Searle explicitly identifies assertives, directives, commissives, expressives, and declaratives, we introduced interrogatives and indirect speech acts due to their frequent occurrence and distinct pragmatic functions in retrieval-oriented user queries.

Although interrogatives could theoretically be subsumed under directives in Searle’s original taxonomy, we explicitly distinguish them in our method for practical computational convenience, given their unique linguistic markers and high frequency in information retrieval contexts.

Indirect speech acts commonly occur as polite or nuanced formulations of requests or questions, necessitating specific computational treatment.

Specifically, interrogative and indirect speech acts represent specialized cases of directives, explicitly requesting that the recipient resolves the speaker’s informational need either directly or indirectly. Their distinct linguistic markers significantly impact propositional content extraction in retrieval tasks

While interrogatives could theoretically be grouped under directives in Searle’s taxonomy, practical experience reveals clear pragmatic distinctions between direct questions (e.g., “What is the capital of France?”) and polite or indirect requests (e.g., “Could you tell me the capital of France?”). These variations carry unique linguistic markers and justify separate treatment. Including these additional categories allows our methodology to more effectively address the diverse forms of user queries encountered in real-world retrieval scenarios.

As shown in Table \ref{tab:speechact}, assertive utterances typically requires no alteration or only minimal adjustments since their propositional content is already explicitly stated. However, interrogative forms are converted into affirmative statements by removing question markers. Requests or directives, particularly those framed politely, are simplified to nominal or topical phrases capturing the essence of the inquiry. Expressive speech acts are distilled by omitting subjective or emotional expressions, thereby isolating factual content relevant for retrieval. Commissive acts, involving commitments about future actions. Indirect speech act sare also simplified into direct propositions by removing introductory clauses.

\end{multicols}
\begin{table}
\caption{Examples of Speech Act Types and Extracted Propositional Content}
\label{tab:speechact}
\small
\begin{tabular}{p{0.14\textwidth} p{0.39\textwidth} p{0.39\textwidth}}
\toprule
\textbf{Speech Act Type} & \textbf{User Utterance} & \textbf{Extracted Propositional Content} \\
\midrule
\textit{Assertive} & The new Anatel Licensing Regulation began to be enforced on November 3. & The new Anatel Licensing Regulation began to be enforced on November 3. \\
\textit{Interrogative} & Has TelComp challenged the charging of TPU fees in other municipalities? & TelComp challenged the charging of TPU fees in other municipalities. \\
\textit{Directive} & List the discounts Disney+ offered in its 2020 pre-sale in Brazil. & Disney+ 2020 pre-sale discounts in Brazil \\
\textit{Expressive} & It’s surprising that Seaborn quickly activated its services after connecting the AMX-1 cable. & Seaborn quickly activated its services after connecting the AMX-1 cable. \\
\textit{Commissive} & I will share my summary of the 5x5 TECSummit after I read all related articles. & Sharing summary of the 5x5 TECSummit after reading all related articles. \\
\textit{Indirect} & Could it be that Seaborn’s partnership with Telecall ensures better redundancy in Rio? & Seaborn's partnership with Telecall ensures better redundancy in Rio. \\
\textit{Declarative} & We hereby announce our intention to use the 5x5 TECSummit findings in our corporate strategy. & Use of the 5x5 TECSummit findings in our corporate strategy. \\
\bottomrule
\end{tabular}
\end{table}
\begin{multicols}{2}\raggedcolumns

Our propositional content extraction process leverages GPT-4 to dynamically identify and transform illocutionary force markers. Some of the linguistic cues addressed by our system include:\footnote{The propositional content extraction method presented in this study assumes a corpus predominantly composed of assertive statements. However, if the knowledge base primarily consists of interrogative content, the transformation process should be adjusted accordingly, either preserving or converting all arguments, including assertive statements, into interrogative forms.}

\begin{itemize}
\item \textbf{Question Indicators:} Sentences ending in question marks, beginning with interrogatives (``what,'' ``where,'' ``how,'' etc.), or auxiliary verbs (``is,'' ``does''). Questions are typically converted into declarative forms or topical noun phrases, depending on context.
\item \textbf{Directive/Imperative Markers:} Politeness terms such as ``please'' or verbs commanding action (``show,'' ``tell,'' ``provide'') are removed, resulting in concise phrases highlighting the query’s central topic.
\item \textbf{Performative Verbs:} Verbs explicitly stating the speaker's intent (``I ask,'' ``I request,'' ``I wonder'') are identified and eliminated, leaving behind the central proposition.
\item \textbf{Expressive Terms:} Emotional or attitudinal markers (``I'm happy that,'' ``unfortunately'') are removed to retain factual statements that can directly match document content.
\item \textbf{Meta-Conversational Phrases:} General conversational expressions (``can you,'' ``do you know'') are stripped away, leaving content-focused queries.
\end{itemize}

After applying these linguistic transformations, the resulting query fragments or sentences succinctly encapsulate the user’s original information needs. 

\begin{figure}[H]
\centering
\includegraphics[width=0.9\textwidth]{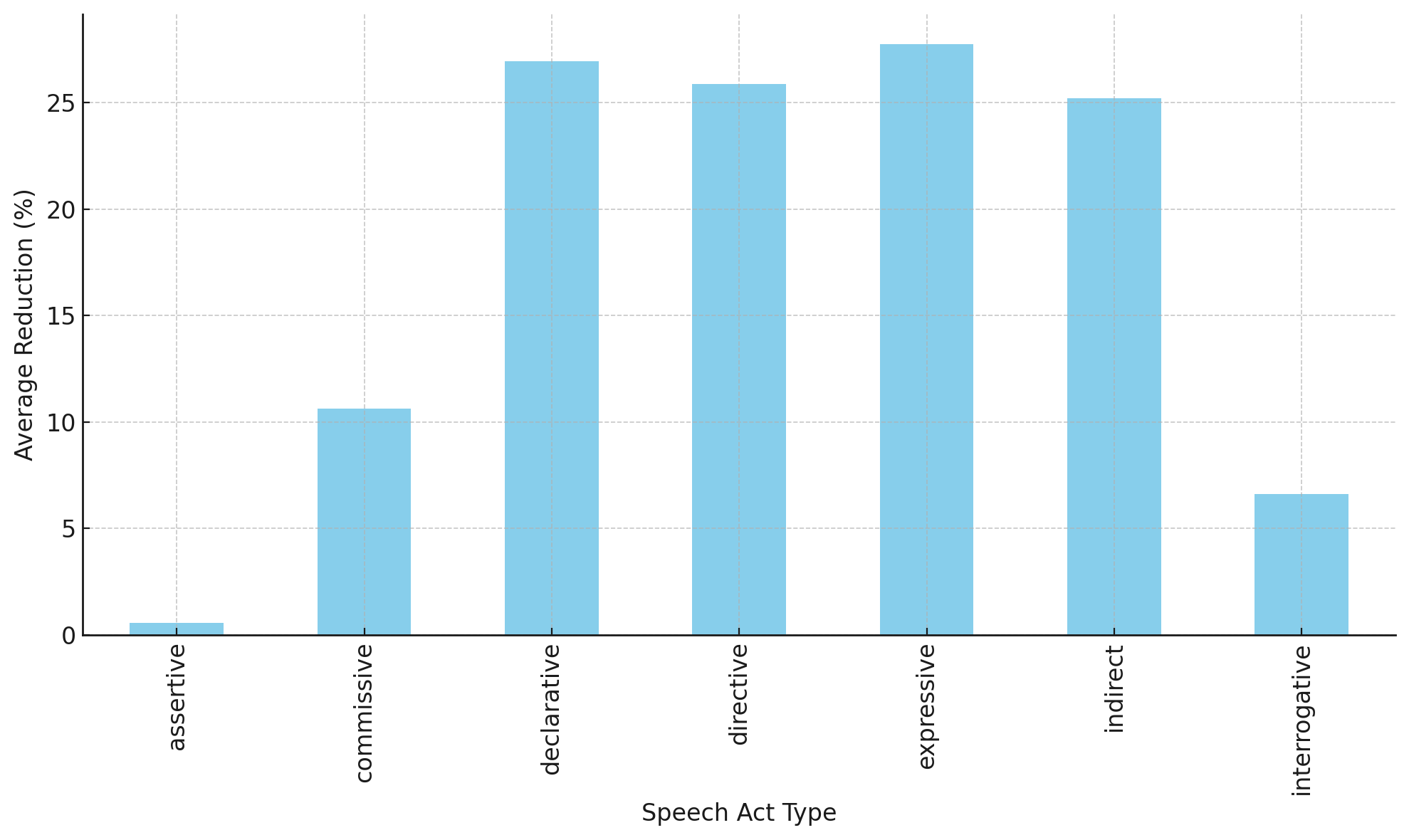}
\caption{Average Character Reduction per Speech Act Type after Propositional Content Extraction}
\label{fig:characters}
\end{figure}

The bar chart in Figure \ref{fig:characters} shows the average percentage reduction in query length, measured by the number of characters, following the propositional content extraction process across different speech act types. Notably, the \textit{declarative}, \textit{directive}, \textit{expressive}, and \textit{indirect} speech acts exhibit the greatest average reductions. This reflects the effectiveness of removing illocutionary markers to yield shorter, more concise formulations that better align with retrieval-oriented embeddings. Conversely, \textit{assertive} speech acts showed a minimal  reduction, as these queries already tend to clearly state propositional content without substantial pragmatic markers.

\end{multicols}
\begin{multicols}{2}\raggedcolumns

Notably, not all utterances inherently contain clear propositional content relevant for retrieval; speech acts such as greetings or gratitude expressions typically do not provide retrievable content and thus fall outside the scope of our methodology. Our experiments primarily focus on utterances clearly expressing information requests or inquiries, where propositional content extraction significantly aids document retrieval performance.

\subsection{Experimental Setup}

To evaluate the effectiveness of extracting propositional content from user queries, we established a hybrid experimental framework that integrates simulated dialogue interactions with document retrieval from an extensive corpus. This approach allows us to directly assess how the propositional content extraction influences retrieval outcomes.

\begin{itemize}
\item \textbf{Dataset (Knowledge Base):} Our dataset consists of a comprehensive collection of 64,983 news articles published between 1998 and 2025, all related to the Brazilian telecommunications industry. These documents cover diverse topics, including mobile service launches, regulatory updates, financial announcements, and network outages. Following the multi-layered embedding-based retrieval technique proposed by \cite{lima2024unlockinglegalknowledgemultilayered}, each news article was segmented into individual paragraphs, resulting in 189,585 distinct textual passages. Embeddings were generated separately for both entire articles and individual paragraphs, employing OpenAI's \texttt{text-embedding-3-large} model, which produces 3,072-dimensional vectors. In this study, we truncated these embeddings to 256 dimensions using the Matryoshka Representation Learning technique\cite{kusupati2024matryoshkarepresentationlearning}. Our knowledge base comprises a total of 254,568 embedding vectors, effectively capturing semantic nuances and hierarchical relationships across multiple layers within the corpus, thus providing granular retrieval capability aligned with the varying informational needs of users.

\item \textbf{User Interactions (Queries):} We prepared a set of 63 user queries (9 queries for each speech act type) that reflect typical information-seeking scenarios within telecom contexts. These queries include a mix of direct and indirect questions, polite requests, and assertive statements to closely simulate actual user interactions.

\item \textbf{Retrieval Methods:} We employed two retrieval methodologies to generate embeddings from the queries:
\begin{enumerate}
    \item \textbf{Baseline Method:} This conventional method involved embedding the original user queries directly,  using the OpenAI's   \texttt{text-embedding-3-large} model. All aspects of the original query—including interrogative phrasing, politeness terms, and illocutionary markers—contributed to the generated vectors.

    \item \textbf{Propositional Method (Our Approach):} In this method, each user query underwent preprocessing to explicitly isolate and extract its propositional content. Specifically, illocutionary indicators, question formats, and polite expressions were systematically removed or transformed into declarative statements or concise topical phrases. These simplified propositional phrases were then embedded into vectors using the same embedding model. The complete prompt used for propositional content extraction is provided in full at the beginning of the Appendix.

\end{enumerate}
Retrieval results were obtained by computing the cosine similarity between the query vectors and the extensive database of news textual embeddings (full content and paragraphs), with the top-ranked chunks identified accordingly.

\item \textbf{Evaluation Approach:} Given the substantial size of our knowledge base (over 250,000 embedding vectors), traditional manual evaluation of retrieval metrics such as precision and recall was impractical. Instead, our evaluation focused on quantitative comparisons of semantic similarity metrics—minimum, maximum, mean, and standard deviation of cosine similarity scores.

\end{itemize}

This adapted experimental setup enables a meaningful qualitative comparison, demonstrating whether propositional content extraction significantly enhances the relevance of retrieved content in realistic information-seeking scenarios.

\subsection{Example of the Process}

To illustrate how our propositional content extraction method operates in practice, consider the initial user query: \textit{``What did Artur Coimbra say?''} Through our method, guided by speech act theory, this interrogative form is simplified into the assertive proposition: \textit{``Artur Coimbra said.''} By removing interrogative markers such as \textit{``What did,''} the propositional version aligns more effectively with the assertive style prevalent in the corpus, which primarily consists of assertive statements.

The difference between the original query and its propositional form significantly impacts retrieval performance due to the layered nature of our news corpus. Specifically, our telecommunications news dataset employs a multilayered embedding structure, consisting of vectors computed separately for both entire articles and individual paragraphs. Because statements attributed to specific individuals are predominantly found within paragraph-level embeddings, converting queries into propositional forms that emphasize such assertions naturally enhances semantic alignment with these paragraph vectors.

Quantitatively, embedding the original interrogative query retrieved 20 chunks (19 paragraphs and 1 full text), with cosine similarity scores ranging from 0.5298 to 0.5006, containing only 6 occurrences of the core term ``Artur Coimbra.'' In contrast, embedding the propositional query yielded 25 chunks (24 paragraphs and 1 full text), achieving notably higher similarity scores ranging from 0.5491 to 0.5129, with a significantly increased frequency of 21 occurrences of the core term. This improvement in both the quantity of retrieved content and the frequency of key terminology\footnote{This study does not examine the correlation between the frequency of occurrences of the core proposition term and the average similarity scores. However, a preliminary analysis revealed a moderate positive correlation (r = 0.56).} underscores how propositional simplification effectively aligns query embeddings with those of relevant document content.

\section{Results and Discussion}

Given the primary focus of this research was to investigate improvements specifically in semantic similarity---essential for retrieval alignment---we intentionally constrained our evaluation to these similarity metrics. While these findings  support the propositional content extraction approach, future research could further validate them via human-based assessments, user studies on generated answers, or standardized question-answering benchmarks. Additionally, an exploratory correlation analysis between the number of retrieved segments (chunks) and their average semantic similarity revealed an extremely weak positive correlation (Pearson $r = 0.07$). This negligible relationship suggests that retrieving more segments does not inherently improve similarity scores. Because of its minimal explanatory power relative to our core objectives, we opted not to investigate this correlation further.

\begin{figure}[H]
\centering
\includegraphics[width=\textwidth]{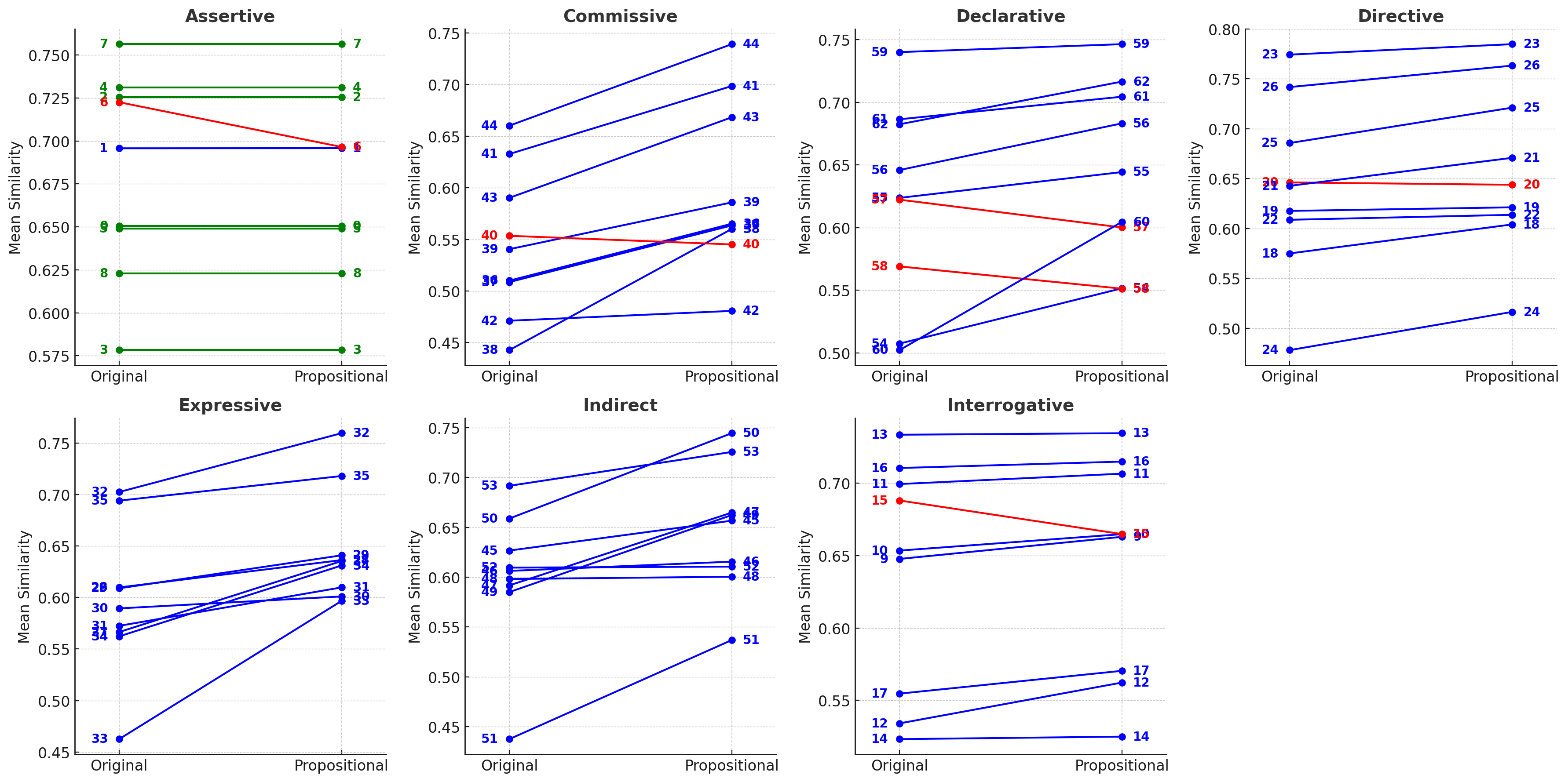}
\caption{Comparison of Mean Similarity Between Original and Propositional Query Forms by Speech Act Type and Query Number. Colors indicate change: green = no change; blue = improved similarity; red = decreased similarity.}
\label{fig:slope}
\end{figure}

\begin{figure}[H]
\centering
\includegraphics[width=\textwidth]{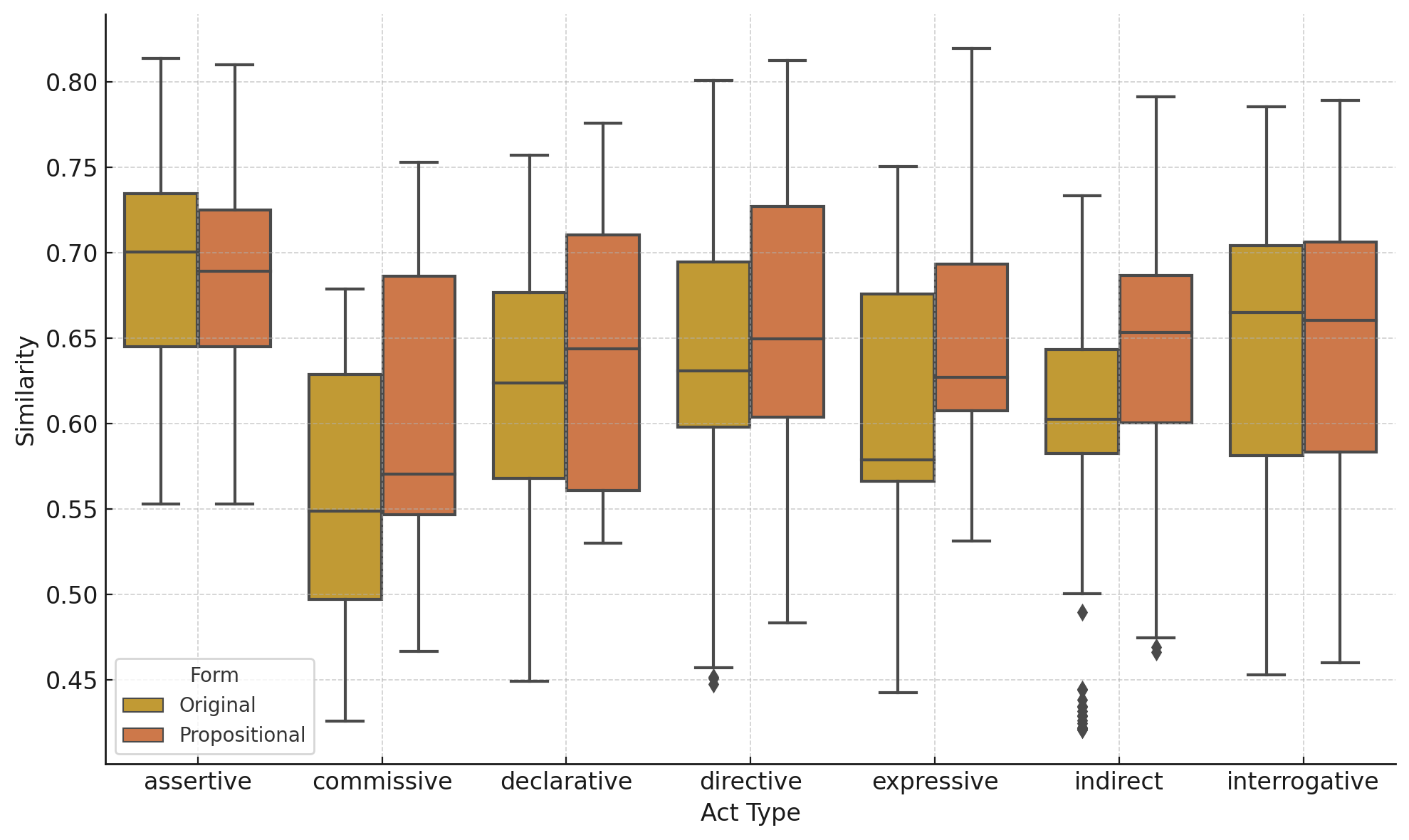}
\caption{Comparison of Similarity Distributions Between Original and Propositional Query Forms by Speech Act Type}
\label{fig:similarity}
\end{figure}

The results obtained from our propositional content extraction method provide compelling evidence of its effectiveness in improving retrieval performance within RAG frameworks. Figures~\ref{fig:slope} and~\ref{fig:similarity} clearly illustrate how similarity distributions differ between the original and propositional forms of queries across multiple speech act types. 

Table~\ref{tab:aggregated_results} summarizes the primary similarity metrics (minimum, maximum, and mean) for each speech act category. By focusing on these measures, we explicitly capture whether propositional content extraction leads to enhanced alignment between queries and the document embeddings. Across the board, our findings highlight notable benefits for nearly all speech act types, particularly for maximum and mean similarity scores.

\end{multicols}

\begin{multicols}{2}\raggedcolumns

\end{multicols}

\begin{table}[H]
\centering
\caption{Similarity Metrics by Speech Act Type}
\label{tab:aggregated_results}
\small
\begin{tabular}{lcccccc}
\toprule
Speech Act & \multicolumn{2}{c}{Min Similarity} & \multicolumn{2}{c}{Max Similarity} & \multicolumn{2}{c}{Mean Similarity} \\
 & Original & Propositional & Original & Propositional & Original & Propositional \\
\midrule
Assertive & \textbf{0.5531} & \textbf{0.5531} & \textbf{0.8137} & 0.8100 & \textbf{0.6871} & 0.6827 \\
Commissive & 0.4261 & \textbf{0.4667} & 0.6786 & \textbf{0.7530} & 0.5571 & \textbf{0.6001} \\
Declarative & 0.4493 & \textbf{0.5300} & 0.7570 & \textbf{0.7761} & 0.6224 & \textbf{0.6396} \\
Directive & 0.4478 & \textbf{0.4836} & 0.8008 & \textbf{0.8124} & 0.6405 & \textbf{0.6619} \\
Expressive & 0.4429 & \textbf{0.5315} & 0.7504 & \textbf{0.8196} & 0.6039 & \textbf{0.6482} \\
Indirect & 0.4209 & \textbf{0.4666} & 0.7334 & \textbf{0.7913} & 0.6002 & \textbf{0.6503} \\
Interrogative & 0.4532 & \textbf{0.4601} & 0.7854 & \textbf{0.7891} & 0.6449 & \textbf{0.6526} \\
\bottomrule
\end{tabular}%
\end{table}

\begin{multicols}{2}\raggedcolumns

As initially predicted, the \textit{assertive} category exhibited negligible differences in similarity metrics before and after propositional transformation. This result reaffirms that assertive utterances typically embed propositional content clearly and therefore see minimal gains from removing illocutionary markers.

By contrast, larger improvements emerged when examining certain categories in terms of \emph{maximum similarity}. The \textit{commissive} speech acts saw a substantial increase in maximum similarity (from $0.6786$ to $0.7530$), suggesting that stripping future-oriented or promissory language clarified the core information. Similarly, \textit{expressive} speech acts benefited from the removal of emotional or subjective elements: their maximum similarity jumped from $0.7504$ to $0.8196$, reflecting how emotional tone can obscure pertinent factual content.

Meanwhile, \textit{indirect} speech acts demonstrated a noticeable lift in maximum similarity, from $0.7334$ to $0.7913$. This underscores that explicitly reframing an implicit or politely worded request into a direct statement can considerably enhance semantic alignment with stored documents. In a similar vein, \textit{interrogative} and \textit{directive} speech acts benefited from converting question or command forms into concise statements or noun phrases, which typically align more naturally with the assertive style found in most knowledge-base texts. Consequently, interrogative queries advanced from a maximum similarity of $0.7854$ to $0.7891$, and directives improved from $0.8008$ to $0.8124$.

Notably, \textit{declarative} acts improved from $0.7570$ to $0.7761$ in maximum similarity, albeit less dramatically than other categories. This outcome aligns with expectations, given the already formal and statement-based nature of assertives in the corpus.

Altogether, these findings demonstrate that propositional content extraction significantly refines embedding-based retrieval across most speech act types, particularly concerning maximum similarity. The performance variations align with our premise that utterances rich in illocutionary or pragmatic markers benefit the most from this reformulation.

While these results validate propositional content extraction as a practical approach to mitigate embedding mismatches, several considerations remain. First, our experimental methodology does not utilize standard precision or recall metrics, as the corpus size rendered manual assessment of top-$k$ retrieved chunks impractical. Future evaluations may therefore benefit from more detailed, human-centered reviews or standardized IR metrics after domain experts establish relevant ground-truth labels.

Overall, our experiments underscore the importance of explicitly addressing propositional content within user queries to enhance semantic alignment in retrieval tasks. This highlights the broader potential of applying speech act theory to computational contexts, showing that explicitly managing pragmatic elements can tangibly improve embedding-based retrieval pipelines. Future research extending this approach to larger and more diverse corpora, or combining it with additional query reformulation strategies, may offer deeper insights into the interplay between linguistic pragmatics and neural retrieval models.

Further quantitative details are provided in the Appendix, Tables section.

\section{Conclusion}

In this study, we examined the impact of extracting propositional content from user queries on improving retrieval performance within RAG systems. Drawing on the foundational work of Austin and Searle, we proposed a practical methodology for systematically converting user queries into simplified propositional equivalents, effectively removing illocutionary markers. Our empirical evaluation—based on 63 user queries spanning seven speech-act categories within a specialized Brazilian telecommunications news corpus—showed clear retrieval benefits when queries were reformulated to emphasize their core propositional content.

The results validate our hypothesis that focusing on propositional information strengthens embedding alignment between user queries and indexed document embeddings. Notably, interrogative, indirect, expressive, and commissive speech acts exhibited significant gains, underscoring that clarity of propositional content can help mitigate the embedding mismatches often observed in RAG frameworks.

Despite these positive outcomes, several limitations should be acknowledged. First, due to the large size of our corpus, we relied on a qualitative evaluation method, which may constrain broader generalization. Second, this study did not employ widely used public datasets—such as MS MARCO, TREC, or TQA—but instead used queries tailored to the Brazilian telecommunications news corpus. While our findings demonstrate improvements in semantic similarity, they pertain specifically to alignment within the embedding space rather than overall retrieval effectiveness. Hence, further work is necessary to incorporate precision, recall, and user-based relevance assessments for a more comprehensive evaluation of practical efficacy.

Our preliminary investigations also revealed that reducing query length must be approached cautiously. For instance, selectively removing a nonessential modifier increased maximum similarity (e.g., from $0.7444$ to $0.7866$), whereas omitting key phrases led to a drop (down to $0.7334$). This highlights the need to balance brevity and semantic preservation, suggesting a promising direction for deeper inquiry into how best to optimize this trade-off.

Future research might expand these findings to broader domains and multilingual contexts, thereby assessing the proposed approach’s robustness and generalizability. Additionally, incorporating human-in-the-loop validation strategies could offer valuable insights into user satisfaction and the effectiveness of query reformulation in real-world applications. In conclusion, these findings underscore the substantial potential of integrating propositional content extraction, grounded in speech act theory, into embedding-based retrieval pipelines—particularly within RAG systems—to address the challenges posed by pragmatic markers and to improve retrieval performance.

In conclusion, our findings underscore the significant potential of integrating propositional content extraction grounded in speech act theory within embedding-based retrieval systems, particularly RAG systems.

\bibliographystyle{unsrt}
\bibliography{RAGPropContent}

\begin{thebibliography}{10}

\bibitem{austin1962}
J.~L. Austin.
\newblock {\em How to Do Things with Words}.
\newblock Oxford University Press, Oxford, UK, 1962.

\bibitem{searle1969}
John~R. Searle.
\newblock {\em Speech Acts: An Essay in the Philosophy of Language}.
\newblock Cambridge University Press, Cambridge, UK, 1969.

\bibitem{searle1979}
John~R. Searle.
\newblock {\em Expression and Meaning: Studies in the Theory of Speech Acts}.
\newblock Cambridge University Press, Cambridge, UK, 1979.

\bibitem{lewis2020}
Patrick Lewis, Ethan Perez, Aleksandra Piktus, Fabio Petroni, Vladimir Karpukhin, Naman Goyal, Heinrich K{\"u}ttler, Mike Lewis, Wen-tau Yih, Tim Rockt{\"a}schel, Sebastian Riedel, and Douwe Kiela.
\newblock Retrieval-augmented generation for knowledge-intensive nlp tasks.
\newblock In {\em Advances in Neural Information Processing Systems (NeurIPS)}, 2020.

\bibitem{karpukhin2020}
Vladimir Karpukhin, Barlas Oguz, Sewon Min, Patrick Lewis, Ledell Wu, Sergey Edunov, Danqi Chen, and Wen-tau Yih.
\newblock Dense passage retrieval for open-domain question answering.
\newblock In {\em Proceedings of the 2020 Conference on Empirical Methods in Natural Language Processing (EMNLP)}, pages 6769--6781, 2020.

\bibitem{gao2022}
Leo Gao, Aman Madaan, Shuyan Zhou, Uri Alon, Pengfei Liu, Yiming Yang, Jamie Callan, and Graham Neubig.
\newblock Precise zero-shot dense retrieval without relevance labels.
\newblock {\em arXiv preprint arXiv:2212.10496}, 2022.

\bibitem{SearleVanderveken1985}
John~R. Searle and Daniel Vanderveken.
\newblock {\em Foundations of Illocutionary Logic}.
\newblock Cambridge University Press, Cambridge, UK, 1985.

\bibitem{Stolcke2000}
Andreas Stolcke, Elizabeth Shriberg, Rebecca Bates, Noah Coccaro, Daniel Jurafsky, Rachel Martin, Marie Meteer, Klaus Ries, Paul Taylor, and Carol~Van Ess-Dykema.
\newblock Dialogue act modeling for automatic tagging and recognition of conversational speech.
\newblock {\em Computational Linguistics}, 26(3):339--373, 2000.

\bibitem{Jurafsky2009}
Dan Jurafsky and James~H. Martin.
\newblock {\em Speech and Language Processing: An Introduction to Natural Language Processing, Computational Linguistics, and Speech Recognition}.
\newblock Pearson Education, 2009.

\bibitem{Ma2021}
Xuezhe Ma, Pengcheng Xu, Jinlan Wang, Yu~Chen, Zhiguo Huang, Pengfei Zhang, and Hai Zhao.
\newblock Improving question answering with explicitly generated questions.
\newblock In {\em ACL 2021}, 2021.

\bibitem{Reimers2019}
Nils Reimers and Iryna Gurevych.
\newblock Sentence-bert: Sentence embeddings using siamese bert-networks.
\newblock In {\em Proceedings of the 2019 Conference on Empirical Methods in Natural Language Processing and the 9th International Joint Conference on Natural Language Processing (EMNLP-IJCNLP)}, pages 3982--3992, 2019.

\bibitem{lima2024unlockinglegalknowledgemultilayered}
João~Alberto de~Oliveira~Lima.
\newblock Unlocking legal knowledge with multi-layered embedding-based retrieval.
\newblock {\em arXiv preprint arXiv:2411.07739}, 2024.

\bibitem{kusupati2024matryoshkarepresentationlearning}
Aditya Kusupati, Gantavya Bhatt, Aniket Rege, Matthew Wallingford, Aditya Sinha, Vivek Ramanujan, William Howard-Snyder, Kaifeng Chen, Sham Kakade, Prateek Jain, and Ali Farhadi.
\newblock Matryoshka representation learning.
\newblock {\em arXiv preprint arXiv:2205.13147}, 2024.

\end{thebibliography}

\section*{Acknowledgments}
The author is grateful to Ari Hershowitz for his comments on the draft of this paper, which have contributed to its improvement. 

\end{multicols}

\clearpage

\hypertarget{appendix-questions}{%
\section{Appendix}\label{appendix-questions}}

\subsection*{System Prompt for Propositional Content Extraction}
\begin{verbatim}
You are an assistant specialized in extracting propositional content from user queries 
based on Speech Act Theory.

Your task is to transform user inputs into simplified statements that clearly preserve 
the core propositional content, systematically removing linguistic indicators of 
illocutionary force to optimize retrieval performance.

Apply these enhanced transformation rules for each speech act category:
1. Assertives:
   - Preserve the original content and phrasing exactly as provided, without alterations.
2. Interrogatives:
   - Convert questions into clear, direct affirmative statements.
   - Completely remove question markers ("?"), interrogative words 
   ("what," "who," "where," "when," "why," "how"), and auxiliary verbs 
   in questions ("is," "does," "did," "can," "will").
3. Directives (requests/commands):
   - Convert commands or requests into concise noun phrases or topical expressions.
   - Eliminate imperative verbs ("show," "provide," "tell") and politeness terms 
   ("please," "kindly").
4. Expressives:
   - Remove all subjective, emotional, or attitudinal markers ("I'm happy," 
   "unfortunately," "luckily"), maintaining strictly factual content.
5. Commissives (speaker commitments/promises):
   - Simplify to reflect the committed action clearly and concisely, omitting 
   explicit performative verbs ("I promise," "I commit," "I will").
   - Express the propositional core as a neutral statement of intended action or 
   future occurrence.
6. Indirect Speech Acts:
   - Eliminate introductory clauses or indirect phrasing (e.g., "I wonder if," 
   "Could you tell me," "Do you know if"), converting indirect queries into direct 
   affirmative statements.
7. Declaratives:
   - Remove introductory declarative phrases explicitly stating the act itself, 
   such as "I declare," "We declare," "I hereby confirm," "I officially proclaim," 
   leaving only the core propositional content clearly expressed.

Specifically target and address these linguistic indicators:
- Question markers: Completely remove punctuation and interrogative terms associated 
with questions.
- Imperative markers: Eliminate command verbs and polite expressions entirely.
- Performative verbs: Omit verbs explicitly declaring intent or commitment 
("I ask," "I request," "I suggest," "I wonder," "I promise," "I commit," "I declare," 
"I hereby confirm," "I officially proclaim").
- Expressive terms: Fully exclude emotional or attitudinal expressions.
- Meta-conversational phrases: Completely omit conversational fillers and 
indirect discourse markers ("can you," "could you," "would you," "do you know," 
"I'd like to know").

Respond ONLY with the extracted propositional content. 
Do NOT include explanations or additional text.
\end{verbatim}

\clearpage

\subsection{Presentation and Interpretation of Query Metrics}

The data is systematically presented in tables organized by speech-act type, followed by query identifier number (QID), and query type (original versus propositional content). Each query is displayed in two distinct rows: the first row shows metrics related to the original query text, and the second row presents metrics derived from its propositional content. Statistical metrics include similarity rates—specifically, the minimum (Min), maximum (Max), mean, and standard deviation (Std)—as well as segment-level statistics, namely the total number of segments (\textit{Segments}) identified and the count of segments unique to each query type (\textit{Dist.}).

Explicitly, column labels represent the following: 

\textbf{QID} (Query Identifier) uniquely identifies each query instance; 

\textbf{Query Text} contains either the original textual form or the propositional representation of the query; 

\textbf{Min}, \textbf{Max}, \textbf{Mean}, and \textbf{Std} indicate, respectively, the lowest, highest, average, and variability measures of semantic similarity scores computed for each query type; 

\textbf{Segments} reports the total number of semantic segments (chunks) found within each query; 

\textbf{Dist.} (Distinct) indicates the count of segments exclusive to either the original or propositional versions of the query, highlighting semantic differentiation between the two versions.

\begin{table}[htbp]
\centering
\caption{Metrics for Assertive queries}
\begin{tabularx}{\textwidth}{|c|X|r|r|r|r|r|r|}
\hline
\textbf{QID} & \textbf{Query Text} &  \textbf{Min} & \textbf{Max} & \textbf{Mean} & \textbf{Std} & \textbf{Segments} \ & \textbf{Dist.} \\
\hline
\multirow{2}{*}{0} & The 5x5 TECSummit took place online from December 7 to December 11. & 0.6136 & 0.7258 & 0.6506 & 0.0337 & 18 & 0 \\
 & The 5x5 TECSummit took place online from December 7 to December 11. & 0.6136 & 0.7258 & 0.6506 & 0.0337 & 18 & 0 \\
\hline
\multirow{2}{*}{1} & Seaborn partnered with Telecall to connect international submarine cables to Rio de Janeiro’s terrestrial infrastructure. & 0.6661 & 0.7914 & 0.6957 & 0.0373 & 26 & 0 \\
 & Seaborn partnered with Telecall to connect international submarine cables to Rio de Janeiro’s terrestrial infrastructure. & 0.6664 & 0.7915 & 0.6958 & 0.0373 & 26 & 0 \\
\hline
\multirow{2}{*}{2} & Abrint actively took part in the OECD interviews about the Brazilian telecom sector. & 0.6991 & 0.8055 & 0.7256 & 0.0336 & 14 & 0 \\
 & Abrint actively took part in the OECD interviews about the Brazilian telecom sector. & 0.6991 & 0.8055 & 0.7256 & 0.0336 & 14 & 0 \\
\hline
\multirow{2}{*}{3} & TelComp won legal cases against municipalities charging the TPU fee. & 0.5547 & 0.6460 & 0.5783 & 0.0294 & 18 & 0 \\
 & TelComp won legal cases against municipalities charging the TPU fee. & 0.5547 & 0.6460 & 0.5783 & 0.0294 & 18 & 0 \\
\hline
\multirow{2}{*}{4} & The new Anatel Licensing Regulation began to be enforced on November 3. & 0.7175 & 0.7525 & 0.7312 & 0.0128 & 11 & 0 \\
 & The new Anatel Licensing Regulation began to be enforced on November 3. & 0.7175 & 0.7525 & 0.7312 & 0.0128 & 11 & 0 \\
\hline
\multirow{2}{*}{5} & Highline appointed Christiano Morette as its new COO. & 0.5531 & 0.7539 & 0.6490 & 0.0835 & 7 & 0 \\
 & Highline appointed Christiano Morette as its new COO. & 0.5531 & 0.7539 & 0.6490 & 0.0835 & 7 & 0 \\
\hline
\multirow{2}{*}{6} & Telefónica’s cybersecurity activities are now under Telefónica Tech in Brazil. & 0.6877 & 0.8137 & 0.7225 & 0.0298 & 25 & 1 \\
 & Telefónica’s cybersecurity activities are under Telefónica Tech in Brazil. & 0.6565 & 0.7792 & 0.6965 & 0.0276 & 25 & 1 \\
\hline
\multirow{2}{*}{7} & The Anatel postponed a satellite rights auction originally set for November 8, 2006, to November 17, 2006. & 0.7252 & 0.8100 & 0.7563 & 0.0262 & 20 & 0 \\
 & The Anatel postponed a satellite rights auction originally set for November 8, 2006, to November 17, 2006. & 0.7252 & 0.8100 & 0.7563 & 0.0262 & 20 & 0 \\
\hline
\multirow{2}{*}{8} & Blob sold over 1,200 location-enabled phones with Claro chips in one month. & 0.5958 & 0.7439 & 0.6230 & 0.0470 & 10 & 0 \\
 & Blob sold over 1,200 location-enabled phones with Claro chips in one month. & 0.5958 & 0.7439 & 0.6230 & 0.0470 & 10 & 0 \\
\hline
\end{tabularx}
\label{tab:assertive}
\end{table}

\begin{table}[htbp]
\centering
\caption{Metrics for Interrogative queries}
\begin{tabularx}{\textwidth}{|c|X|r|r|r|r|r|r|}
\hline
\textbf{QID} & \textbf{Query Text} &  \textbf{Min} & \textbf{Max} & \textbf{Mean} & \textbf{Std} & \textbf{Segments} \ & \textbf{Dist.} \\
\hline
\multirow{2}{*}{9} & Did the 5x5 TECSummit discuss the increasing importance of digital transformation in diverse market segments? & 0.6357 & 0.6657 & 0.6477 & 0.0099 & 15 & 3 \\
 & The 5x5 TECSummit discussed the increasing importance of digital transformation in diverse market segments. & 0.6446 & 0.6981 & 0.6630 & 0.0140 & 23 & 11 \\
\hline
\multirow{2}{*}{10} & Is Seaborn providing international connectivity through the AMX-1 submarine cable? & 0.6050 & 0.7057 & 0.6535 & 0.0320 & 25 & 2 \\
 & Seaborn provides international connectivity through the AMX-1 submarine cable. & 0.6097 & 0.7356 & 0.6648 & 0.0391 & 26 & 3 \\
\hline
\multirow{2}{*}{11} & Did Abrint contribute data to the OECD reports on Brazil’s digital era? & 0.6630 & 0.7772 & 0.6995 & 0.0395 & 10 & 1 \\
 & Abrint contributed data to the OECD reports on Brazil’s digital era. & 0.6775 & 0.7891 & 0.7066 & 0.0310 & 16 & 7 \\
\hline
\multirow{2}{*}{12} & Has TelComp challenged the charging of TPU fees in other municipalities? & 0.5170 & 0.6127 & 0.5340 & 0.0278 & 17 & 13 \\
 & TelComp challenged the charging of TPU fees in other municipalities. & 0.5459 & 0.6163 & 0.5623 & 0.0168 & 28 & 24 \\
\hline
\multirow{2}{*}{13} & Are operators required to harmonize station licensing under the new Anatel RGL? & 0.7221 & 0.7854 & 0.7336 & 0.0143 & 18 & 5 \\
 & Operators are required to harmonize station licensing under the new Anatel RGL. & 0.7186 & 0.7742 & 0.7346 & 0.0157 & 24 & 11 \\
\hline
\multirow{2}{*}{14} & What is Christiano Morette’s role at Highline? & 0.4532 & 0.6020 & 0.5231 & 0.0587 & 7 & 2 \\
 & Christiano Morette's role at Highline. & 0.4601 & 0.5998 & 0.5249 & 0.0581 & 7 & 2 \\
\hline
\multirow{2}{*}{15} & Has Telefónica consolidated its cybersecurity operations under a single business unit? & 0.6638 & 0.7277 & 0.6881 & 0.0206 & 14 & 3 \\
 & Telefónica consolidated its cybersecurity operations under a single business unit. & 0.6470 & 0.7149 & 0.6649 & 0.0232 & 17 & 6 \\
\hline
\multirow{2}{*}{16} & When did the Anatel first postpone the satellite license auction in 2006? & 0.6834 & 0.7487 & 0.7106 & 0.0233 & 18 & 7 \\
 & Anatel first postponed the satellite license auction in 2006. & 0.6916 & 0.7565 & 0.7150 & 0.0187 & 17 & 6 \\
\hline
\multirow{2}{*}{17} & Does Blob’s device come with a built-in location system for customers in Rio and São Paulo? & 0.5395 & 0.5975 & 0.5546 & 0.0160 & 16 & 6 \\
 & Blob's device comes with a built-in location system for customers in Rio and São Paulo. & 0.5547 & 0.6164 & 0.5704 & 0.0163 & 15 & 5 \\
\hline
\end{tabularx}
\label{tab:interrogative}
\end{table}

\begin{table}[htbp]
\centering
\caption{Metrics for Directive queries}
\begin{tabularx}{\textwidth}{|c|X|r|r|r|r|r|r|}
\hline
\textbf{QID} & \textbf{Query Text} &  \textbf{Min} & \textbf{Max} & \textbf{Mean} & \textbf{Std} & \textbf{Segments} \ & \textbf{Dist.} \\
\hline
\multirow{2}{*}{18} & Show me details about the 5x5 TECSummit and its schedule. & 0.5362 & 0.6211 & 0.5750 & 0.0256 & 23 & 2 \\
 & 5x5 TECSummit details and schedule & 0.5727 & 0.6645 & 0.6040 & 0.0259 & 24 & 3 \\
\hline
\multirow{2}{*}{19} & Provide information on how Seaborn and Telecall are improving submarine cable connectivity. & 0.5979 & 0.7034 & 0.6176 & 0.0289 & 21 & 5 \\
 & Seaborn and Telecall improving submarine cable connectivity. & 0.5958 & 0.7014 & 0.6212 & 0.0284 & 23 & 7 \\
\hline
\multirow{2}{*}{20} & List the discounts Disney+ offered in its 2020 pre-sale in Brazil. & 0.6204 & 0.7132 & 0.6461 & 0.0345 & 11 & 4 \\
 & Disney+ 2020 pre-sale discounts in Brazil & 0.6167 & 0.7120 & 0.6438 & 0.0344 & 11 & 4 \\
\hline
\multirow{2}{*}{21} & Explain Abrint’s role in shaping the OECD’s telecom recommendations. & 0.6203 & 0.6794 & 0.6427 & 0.0185 & 17 & 1 \\
 & Abrint's role in shaping the OECD's telecom recommendations. & 0.6480 & 0.7202 & 0.6708 & 0.0209 & 18 & 2 \\
\hline
\multirow{2}{*}{22} & Give me the legal basis of TelComp’s victories regarding TPU fees. & 0.5807 & 0.6580 & 0.6087 & 0.0229 & 20 & 6 \\
 & Legal basis of TelComp’s victories regarding TPU fees. & 0.5836 & 0.6769 & 0.6137 & 0.0270 & 31 & 17 \\
\hline
\multirow{2}{*}{23} & Summarize the key changes in Anatel’s new Licensing Regulation of November 3. & 0.7523 & 0.8008 & 0.7742 & 0.0191 & 13 & 2 \\
 & Key changes in Anatel’s new Licensing Regulation of November 3. & 0.7700 & 0.8124 & 0.7847 & 0.0145 & 15 & 4 \\
\hline
\multirow{2}{*}{24} & Tell me more about the appointment of Christiano Morette at Highline. & 0.4478 & 0.5440 & 0.4781 & 0.0317 & 16 & 5 \\
 & Appointment of Christiano Morette at Highline. & 0.4836 & 0.5851 & 0.5163 & 0.0377 & 14 & 3 \\
\hline
\multirow{2}{*}{25} & Outline how Telefónica Tech is structured in Brazil for digital services. & 0.6703 & 0.7446 & 0.6856 & 0.0191 & 28 & 17 \\
 & Telefónica Tech structure in Brazil for digital services. & 0.6962 & 0.7969 & 0.7211 & 0.0264 & 29 & 18 \\
\hline
\multirow{2}{*}{26} & Describe the reasons for Anatel postponing the satellite auction deadlines. & 0.7269 & 0.7909 & 0.7417 & 0.0164 & 21 & 6 \\
 & Reasons for Anatel postponing the satellite auction deadlines. & 0.7445 & 0.7956 & 0.7632 & 0.0136 & 23 & 8 \\
\hline
\end{tabularx}
\label{tab:directive}
\end{table}

\begin{table}[htbp]
\centering
\caption{Metrics for Expressive queries}
\begin{tabularx}{\textwidth}{|c|X|r|r|r|r|r|r|}
\hline
\textbf{QID} & \textbf{Query Text} &  \textbf{Min} & \textbf{Max} & \textbf{Mean} & \textbf{Std} & \textbf{Segments} \ & \textbf{Dist.} \\
\hline
\multirow{2}{*}{27} & I am thrilled about the 5x5 TECSummit’s focus on digital markets. & 0.5564 & 0.6143 & 0.5664 & 0.0134 & 18 & 5 \\
 & 5x5 TECSummit’s focus on digital markets. & 0.6182 & 0.6599 & 0.6354 & 0.0142 & 23 & 10 \\
\hline
\multirow{2}{*}{28} & It’s surprising that Seaborn quickly activated its services after connecting the AMX-1 cable. & 0.5666 & 0.6767 & 0.6098 & 0.0374 & 22 & 3 \\
 & Seaborn quickly activated its services after connecting the AMX-1 cable. & 0.5947 & 0.7173 & 0.6363 & 0.0386 & 27 & 8 \\
\hline
\multirow{2}{*}{29} & I am excited to see Disney+ offering special combos in Brazil. & 0.5688 & 0.6989 & 0.6089 & 0.0493 & 13 & 3 \\
 & Disney+ offering special combos in Brazil. & 0.6079 & 0.7259 & 0.6409 & 0.0432 & 14 & 4 \\
\hline
\multirow{2}{*}{30} & I am pleased that Abrint’s input influenced international telecom standards. & 0.5704 & 0.6272 & 0.5894 & 0.0170 & 11 & 6 \\
 & Abrint's input influenced international telecom standards. & 0.5814 & 0.6467 & 0.6010 & 0.0173 & 15 & 10 \\
\hline
\multirow{2}{*}{31} & I find it commendable that TelComp defended operators against undue TPU fees. & 0.5590 & 0.5922 & 0.5725 & 0.0099 & 25 & 14 \\
 & TelComp defended operators against undue TPU fees. & 0.6018 & 0.6224 & 0.6098 & 0.0071 & 24 & 13 \\
\hline
\multirow{2}{*}{32} & It’s reassuring to see Anatel simplifying station licensing with the new RGL. & 0.6824 & 0.7504 & 0.7024 & 0.0198 & 13 & 4 \\
 & Anatel simplifying station licensing with the new RGL. & 0.7385 & 0.8196 & 0.7596 & 0.0231 & 14 & 5 \\
\hline
\multirow{2}{*}{33} & I am glad Highline is investing in new leadership with Christiano Morette. & 0.4429 & 0.5606 & 0.4624 & 0.0435 & 7 & 6 \\
 & Highline is investing in new leadership with Christiano Morette. & 0.5315 & 0.6853 & 0.5967 & 0.0704 & 7 & 6 \\
\hline
\multirow{2}{*}{34} & It’s fascinating how Telefónica Tech integrates cybersecurity solutions. & 0.5339 & 0.6171 & 0.5622 & 0.0233 & 18 & 4 \\
 & Telefónica Tech integrates cybersecurity solutions. & 0.5976 & 0.6872 & 0.6310 & 0.0265 & 23 & 9 \\
\hline
\multirow{2}{*}{35} & I am curious about why Anatel delayed the satellite license auction so many times. & 0.6794 & 0.7271 & 0.6940 & 0.0134 & 21 & 10 \\
 & Anatel delayed the satellite license auction multiple times. & 0.7052 & 0.7326 & 0.7180 & 0.0081 & 22 & 11 \\
\hline
\end{tabularx}
\label{tab:expressive}
\end{table}

\begin{table}[htbp]
\centering
\caption{Metrics for Commissive queries}
\begin{tabularx}{\textwidth}{|c|X|r|r|r|r|r|r|}
\hline
\textbf{QID} & \textbf{Query Text} &  \textbf{Min} & \textbf{Max} & \textbf{Mean} & \textbf{Std} & \textbf{Segments} \ & \textbf{Dist.} \\
\hline
\multirow{2}{*}{36} & I will share my summary of the 5x5 TECSummit after I read all related articles. & 0.4860 & 0.5550 & 0.5100 & 0.0225 & 21 & 5 \\
 & Sharing summary of the 5x5 TECSummit after reading all related articles. & 0.5372 & 0.6049 & 0.5651 & 0.0179 & 20 & 4 \\
\hline
\multirow{2}{*}{37} & We promise to compare Seaborn’s submarine cable solution with other providers soon. & 0.4849 & 0.5416 & 0.5086 & 0.0181 & 26 & 4 \\
 & Comparison of Seaborn’s submarine cable solution with other providers soon. & 0.5469 & 0.6039 & 0.5635 & 0.0177 & 24 & 2 \\
\hline
\multirow{2}{*}{38} & I vow to update our internal records once Disney+ subscription figures are published. & 0.4261 & 0.5004 & 0.4426 & 0.0192 & 17 & 12 \\
 & Update of internal records after Disney+ subscription figures are published. & 0.5355 & 0.6016 & 0.5601 & 0.0224 & 8 & 3 \\
\hline
\multirow{2}{*}{39} & I intend to present Abrint’s OECD findings at the next association meeting. & 0.5297 & 0.5576 & 0.5405 & 0.0102 & 15 & 8 \\
 & Presentation of Abrint’s OECD findings at the next association meeting. & 0.5662 & 0.6201 & 0.5858 & 0.0204 & 11 & 4 \\
\hline
\multirow{2}{*}{40} & We pledge to support TelComp’s legal strategies regarding TPU fees going forward. & 0.5360 & 0.5900 & 0.5534 & 0.0158 & 27 & 15 \\
 & Support for TelComp’s legal strategies regarding TPU fees going forward. & 0.5359 & 0.5706 & 0.5451 & 0.0085 & 25 & 13 \\
\hline
\multirow{2}{*}{41} & I plan to report on the new Anatel RGL in our monthly regulatory newsletter. & 0.6226 & 0.6483 & 0.6327 & 0.0067 & 29 & 16 \\
 & Report on the new Anatel RGL in the monthly regulatory newsletter. & 0.6904 & 0.7239 & 0.6986 & 0.0083 & 22 & 9 \\
\hline
\multirow{2}{*}{42} & We will highlight Christiano Morette’s operational initiatives in our next board review. & 0.4603 & 0.4853 & 0.4711 & 0.0085 & 19 & 6 \\
 & We highlight Christiano Morette’s operational initiatives in our next board review. & 0.4667 & 0.5034 & 0.4806 & 0.0106 & 16 & 3 \\
\hline
\multirow{2}{*}{43} & I guarantee to analyze Telefónica Tech’s cybersecurity offerings once the data is consolidated. & 0.5579 & 0.6540 & 0.5902 & 0.0258 & 24 & 12 \\
 & Analysis of Telefónica Tech’s cybersecurity offerings after data consolidation. & 0.6488 & 0.6876 & 0.6683 & 0.0122 & 14 & 2 \\
\hline
\multirow{2}{*}{44} & I commit to following the Anatel satellite auction process and sharing all updates. & 0.6517 & 0.6786 & 0.6602 & 0.0073 & 31 & 25 \\
 & Following the Anatel satellite auction process and sharing all updates. & 0.7262 & 0.7530 & 0.7392 & 0.0084 & 16 & 10 \\
\hline
\end{tabularx}
\label{tab:commissive}
\end{table}

\begin{table}[htbp]
\centering
\caption{Metrics for Indirect queries}
\begin{tabularx}{\textwidth}{|c|X|r|r|r|r|r|r|}
\hline
\textbf{QID} & \textbf{Query Text} &  \textbf{Min} & \textbf{Max} & \textbf{Mean} & \textbf{Std} & \textbf{Segments} \ & \textbf{Dist.} \\
\hline
\multirow{2}{*}{45} & I wonder whether the 5x5 TECSummit provided insights into digital transformation. & 0.6104 & 0.6569 & 0.6266 & 0.0149 & 14 & 2 \\
 & The 5x5 TECSummit provided insights into digital transformation. & 0.6390 & 0.6953 & 0.6566 & 0.0170 & 18 & 6 \\
\hline
\multirow{2}{*}{46} & Could it be that Seaborn’s partnership with Telecall ensures better redundancy in Rio? & 0.5806 & 0.6960 & 0.6062 & 0.0304 & 21 & 3 \\
 & Seaborn's partnership with Telecall ensures better redundancy in Rio. & 0.5805 & 0.7183 & 0.6155 & 0.0349 & 25 & 7 \\
\hline
\multirow{2}{*}{47} & I would like to know if Disney+ pre-launch deals attracted a large Brazilian audience. & 0.5739 & 0.6246 & 0.5917 & 0.0160 & 14 & 6 \\
 & Disney+ pre-launch deals attracted a large Brazilian audience. & 0.6504 & 0.7004 & 0.6648 & 0.0172 & 14 & 6 \\
\hline
\multirow{2}{*}{48} & It would be helpful to see if Abrint’s participation influenced the final OECD reports. & 0.5722 & 0.6319 & 0.5981 & 0.0215 & 9 & 4 \\
 & Abrint's participation influenced the final OECD reports. & 0.5697 & 0.6718 & 0.6004 & 0.0353 & 15 & 10 \\
\hline
\multirow{2}{*}{49} & I’m not sure if TelComp’s legal victories apply to other Brazilian municipalities. & 0.5742 & 0.6026 & 0.5850 & 0.0080 & 21 & 12 \\
 & TelComp's legal victories apply to other Brazilian municipalities. & 0.6519 & 0.6801 & 0.6623 & 0.0082 & 26 & 17 \\
\hline
\multirow{2}{*}{50} & I’d be interested to learn if Anatel’s new RGL merged previous licensing procedures. & 0.6422 & 0.6810 & 0.6588 & 0.0139 & 16 & 6 \\
 & Anatel's new RGL merged previous licensing procedures. & 0.7234 & 0.7913 & 0.7444 & 0.0188 & 14 & 4 \\
\hline
\multirow{2}{*}{51} & I wonder how Christiano Morette’s experience will shape Highline’s operations. & 0.4209 & 0.5006 & 0.4374 & 0.0231 & 17 & 13 \\
 & Christiano Morette’s experience will shape Highline’s operations. & 0.4666 & 0.5949 & 0.5368 & 0.0629 & 7 & 3 \\
\hline
\multirow{2}{*}{52} & Could it be that Telefónica Tech aims to unify all digital solutions under one brand? & 0.5917 & 0.6466 & 0.6095 & 0.0169 & 16 & 8 \\
 & Telefónica Tech aims to unify all digital solutions under one brand. & 0.5883 & 0.6951 & 0.6105 & 0.0270 & 28 & 20 \\
\hline
\multirow{2}{*}{53} & I’d like to find out whether Anatel’s postponed satellite auction had more bidders in 2006. & 0.6766 & 0.7334 & 0.6916 & 0.0179 & 18 & 5 \\
 & Anatel's postponed satellite auction had more bidders in 2006. & 0.7122 & 0.7571 & 0.7256 & 0.0132 & 21 & 8 \\
\hline
\end{tabularx}
\label{tab:indirect}
\end{table}

\begin{table}[htbp]
\centering
\caption{Metrics for Declarative queries}
\begin{tabularx}{\textwidth}{|c|X|r|r|r|r|r|r|}
\hline
\textbf{QID} & \textbf{Query Text} &  \textbf{Min} & \textbf{Max} & \textbf{Mean} & \textbf{Std} & \textbf{Segments} \ & \textbf{Dist.} \\
\hline
\multirow{2}{*}{54} & We hereby announce our intention to use the 5x5 TECSummit findings in our corporate strategy. & 0.4925 & 0.5615 & 0.5073 & 0.0162 & 31 & 11 \\
 & Use of the 5x5 TECSummit findings in our corporate strategy. & 0.5300 & 0.6037 & 0.5515 & 0.0171 & 29 & 9 \\
\hline
\multirow{2}{*}{55} & By this statement, I declare that Seaborn’s cable expansion fosters international connectivity. & 0.6092 & 0.6527 & 0.6237 & 0.0126 & 30 & 10 \\
 & Seaborn’s cable expansion fosters international connectivity. & 0.6254 & 0.6780 & 0.6443 & 0.0155 & 28 & 8 \\
\hline
\multirow{2}{*}{56} & I officially proclaim that Disney+ combos are a milestone in Brazilian streaming services. & 0.6200 & 0.6886 & 0.6459 & 0.0176 & 18 & 6 \\
 & Disney+ combos are a milestone in Brazilian streaming services. & 0.6558 & 0.7363 & 0.6831 & 0.0255 & 17 & 5 \\
\hline
\multirow{2}{*}{57} & We declare the integration of Abrint’s data into our official telecom analysis. & 0.6135 & 0.6527 & 0.6222 & 0.0082 & 29 & 21 \\
 & Integration of Abrint’s data into our official telecom analysis. & 0.5884 & 0.6177 & 0.6000 & 0.0092 & 20 & 12 \\
\hline
\multirow{2}{*}{58} & I hereby confirm that TelComp’s TPU fee victories set an important precedent. & 0.5544 & 0.6012 & 0.5689 & 0.0151 & 25 & 15 \\
 & TelComp’s TPU fee victories set an important precedent. & 0.5369 & 0.5978 & 0.5512 & 0.0159 & 28 & 18 \\
\hline
\multirow{2}{*}{59} & We declare that the new Anatel Licensing Regulation is now a reference for simplification. & 0.7275 & 0.7570 & 0.7400 & 0.0107 & 17 & 5 \\
 & The new Anatel Licensing Regulation is now a reference for simplification. & 0.7355 & 0.7761 & 0.7464 & 0.0114 & 18 & 6 \\
\hline
\multirow{2}{*}{60} & I officially endorse Christiano Morette’s appointment as COO of Highline. & 0.4493 & 0.5845 & 0.5021 & 0.0518 & 7 & 3 \\
 & Christiano Morette's appointment as COO of Highline. & 0.5379 & 0.7087 & 0.6045 & 0.0745 & 7 & 3 \\
\hline
\multirow{2}{*}{61} & We pronounce Telefónica Tech as a consolidated cybersecurity hub in Brazil. & 0.6448 & 0.7477 & 0.6864 & 0.0279 & 26 & 10 \\
 & Telefónica Tech is a consolidated cybersecurity hub in Brazil. & 0.6766 & 0.7535 & 0.7044 & 0.0222 & 21 & 5 \\
\hline
\multirow{2}{*}{62} & By this note, I affirm that Anatel’s satellite auction deferrals influenced investor interest. & 0.6728 & 0.7049 & 0.6824 & 0.0091 & 27 & 16 \\
 & Anatel’s satellite auction deferrals influenced investor interest. & 0.7048 & 0.7393 & 0.7165 & 0.0094 & 25 & 14 \\
\hline
\end{tabularx}
\label{tab:declarative}
\end{table}

\end{document}